\documentclass{iosart2c}
\usepackage[utf8]{inputenc}

\usepackage[T1]{fontenc}
\usepackage{times}%

\usepackage{amsmath}
\usepackage{dcolumn}
\usepackage{graphicx}
\usepackage{url}

\newcolumntype{d}[1]{D{.}{.}{#1}}

\firstpage{1} \lastpage{5} \volume{1} \pubyear{2009}

\begin{document}
\begin{frontmatter}                           

%
\title{Bug Localization from Bug Reports: A Multi-Objective Approach}

\runningtitle{Bug Localization from Bug Reports}

\author{Waleed Ahmad},
\author{Mehtab Kiran Suddle}
and
\author{Maryam Bashir \thanks{Corresponding author. E-mail: maryam.bashir@nu.edu.pk}}
\runningauthor{Ahmad et al.}
\address{FAST School of Computing, \\ National University of Computer and  Emerging Sciences, Lahore, Pakistan}

\begin{abstract}
Bug localization is a labor-intensive task, particularly in large software systems. When abnormal behavior occurs, developers must perform repetitive and time-consuming steps to identify faulty files. Previous studies have mainly focused on single-objective localization methods, many of which are limited to specific programming languages. In addition, relying solely on lexical similarity between source code and bug reports is often insufficient due to the natural language nature of bug descriptions. In this study, we propose a class-level automated multi-objective search-based system to identify and rank potentially buggy classes from bug reports. The main objective is to maximize similarity while minimizing the number of suggested faulty files. The evolutionary optimization algorithm SPEA-2 was applied to six open-source Java projects comprising more than 22,000 bug reports. The proposed approach was evaluated against two widely used algorithms, NSGA-II and MOEA/D. Results indicate that SPEA-2 achieved higher precision and recall than both multi-objective and single-objective baseline methods. The proposed recommender system successfully identified buggy classes or files for 88.5\% of bug reports within the top 10 recommendations and 94\% within the top 20. The effectiveness of the model was further validated on an industrial Android project written in Kotlin, demonstrating its adaptability across programming languages.
\end{abstract}


\end{frontmatter}

\section{Introduction}

Software has become an integral part of modern society, with its widespread use influencing nearly every aspect of daily life, ranging from mobile applications to office software. Today, software is no longer merely a convenience but a necessity, particularly in safety-critical and security-sensitive domains such as nuclear energy, aviation, and healthcare. This increasing dependence on software has resulted in greater system complexity and a higher likelihood of software bugs, which may lead to operational failures. Software bugs, commonly caused by logic or coding errors, manifest when a system produces incorrect outputs or exhibits abnormal behavior during execution \cite{almhana}. When such unexpected behavior arises, developers or users document it in the form of a bug report \cite{lam2017}.

Bug reports provide structured information that assists in identifying and resolving software defects. As software systems continue to scale in size and complexity, the number of reported bugs increases substantially. For instance, repositories such as Mozilla and Eclipse collectively contain more than 215,000 documented bug reports, according to recent datasets. These reports play a crucial role in supporting developers and maintainers by guiding system updates and improving software reliability. However, low-quality bug reports can negatively impact bug localization, making the process more time-consuming and error-prone \cite{almhana}. Manually identifying faulty components from natural language descriptions requires extensive system knowledge, highlighting the need for automated bug localization techniques. This process aims to locate defective code segments in order to enhance system reliability, safety, and performance \cite{mills2020}.

\begin{figure}[ht]
\centerline{\includegraphics[width=80mm,height=70mm]{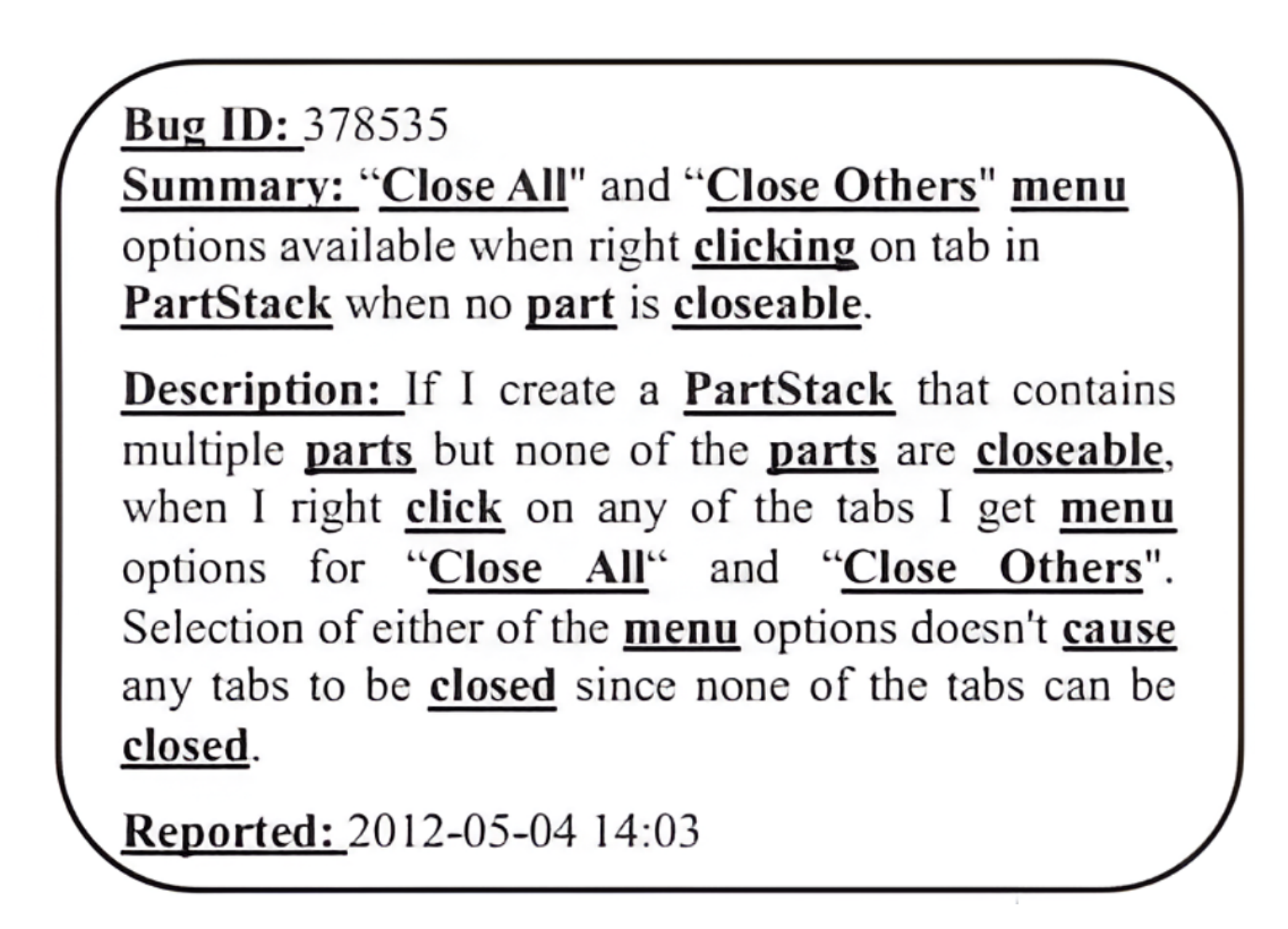}}
\caption{An Eclipse Bug Report Example (ID 378535)}
\label{fig:bug}
\end{figure}

\begin{figure*}
\centerline{\includegraphics[width=110mm,height=70mm]{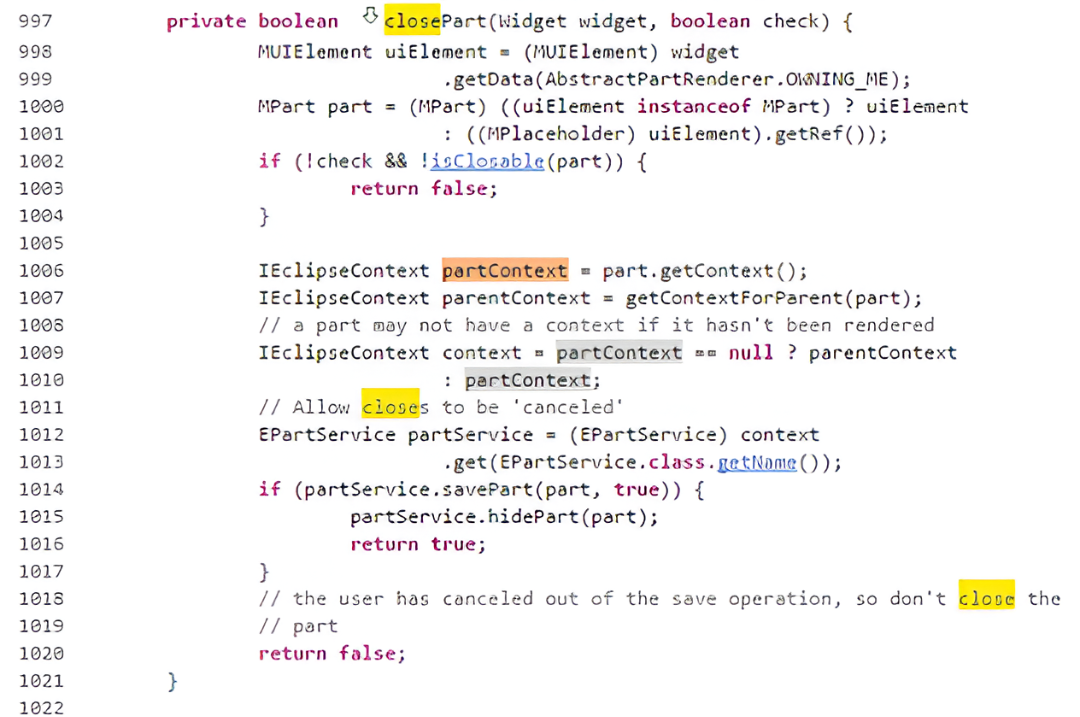}}
\caption{A snippet extracted from the StackRenderer class}
\label{fig:code}
\end{figure*}

\begin{figure*}
\centerline{\includegraphics[width=130mm, height=30mm]{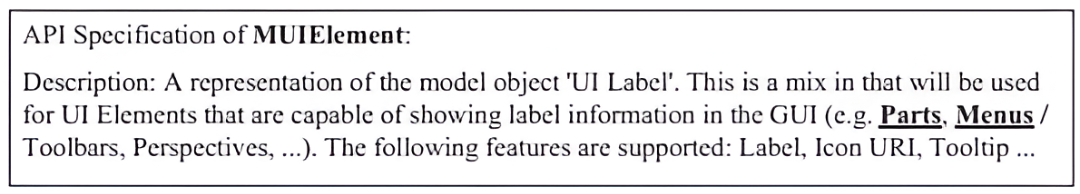}}
\caption{Specification for the API}
\label{fig:api}
\end{figure*}

Effective automated bug localization techniques aim to identify code components that are most likely responsible for reported defects, thereby improving developer productivity by reducing the time and effort required to locate bug sources \cite{almhana2020}. Although many existing approaches employ Text Retrieval (TR) techniques for bug localization, their effectiveness is often limited by the quality of query formulation and evaluation strategies \cite{mills2020}. Furthermore, the linguistic gap between natural language bug reports and programming language source code reduces the effectiveness of purely lexical matching methods \cite{almhana}.

Bug reports are typically written in natural language, whereas source code is expressed using programming languages. As a result, similarity scores between bug reports and source code are generally higher when extensive comments are present or when bug reports explicitly mention code element names. Figure \ref{fig:bug} illustrates an example of a bug report describing an issue related to incorrect menu options for non-closeable parts. The resolution involves examining three related classes extracted from the Eclipse project. As shown in Figure \ref{fig:api}, the StackRenderer class contains a variable named uiElement of type MUIElement. Figure \ref{fig:api} also presents the API specifications of the MUIElement interface, which include terms such as “parts” and “menus” that appear in the bug report description in Figure \ref{fig:code}. This lexical overlap between API documentation and bug reports can contribute to more accurate identification of relevant faulty classes.

This study focuses on automatically ranking files or classes that are most likely to contain defects based on bug report descriptions. The primary objective is to balance reducing the number of ranked files or classes while increasing both history-based and lexical-based similarity using a proposed multi-objective model. The following observations motivate this work:

\begin{itemize}
\item Similar content in previously resolved bug reports may indicate relevance to a new bug report.
\item API documentation is more informative than code comments or element names alone.
\item Recently modified code may still contain defects that lead to abnormal system behavior.
\item The model should prioritize ranking a minimal number of highly relevant classes to reduce time consumption \cite{almhana}.
\end{itemize}

To address these observations, we propose two comprehensive approaches using multi-objective SPEA-2, an enhanced version of the original Strength Pareto Evolutionary Algorithm (SPEA), and a multi-objective evolutionary algorithm based on decomposition (MOEA/D). Additionally, we introduce a novel tokenization technique as a preprocessing step to improve localization accuracy. Our evaluation includes six open-source Java projects \cite{ye2014} as well as an industrial Kotlin-based Android project containing a repository of 100 bug reports.

\subsection{Scope and Objectives}
This research addresses the bug localization problem by formulating it as an optimization task using evolutionary algorithms. Bug reports are collected from six large Java-based projects and one Kotlin-based Android project, with the following objectives:

\begin{enumerate}
\item Formulate bug localization as a multi-objective problem that balances maximizing relevance with minimizing the number of ranked files or classes.
\item Propose a search-based software engineering framework utilizing SPEA-2 and MOEA/D for bug localization.
\item Enhance NSGA-II performance on the selected datasets through a novel tokenization technique.
\item Compare the proposed models with generic lexical similarity, history-based similarity, and NSGA-II approaches.
\item Demonstrate the effectiveness of the proposed framework using bug reports from both Java and Android software projects.
\end{enumerate}

The remainder of this paper is organized as follows: Section \ref{literature} presents a comprehensive literature review, Section \ref{methodology} describes the proposed methodology, Section \ref{eval} outlines the evaluation strategy, and Section \ref{result} discusses the experimental results. Finally, Section \ref{conclusion} concludes the paper and suggests directions for future work.

\begin{center}
\begin{table*}
\footnotesize
\caption{Summary of Literature Review }
\begin{tabular}{|p{2cm} |p{0.7cm}|p{4.5cm}|p{2.5cm}|p{2.5cm}|}
 \hline
Author and Reference & Year & Algorithm & Technique & Dataset \\ [0.5ex] 
 \hline\hline
Saha et al. \cite{saha2013} & 2013 & BLUiR (Bug Localization Using Information Retrieval ) & Information Retrieval & Four Eclipse open source datasets \cite{zhou2012}
\\ \hline

Pandiyan et al. \cite{pandiyan2015} & 2015 & Tabu Search Fault Localization with Path Branch and Bound (TSFL-PBB) & Search-based Software Engineering & 2 UCI datasets
\\ \hline
Rahman et al. \cite{rahman2015} & 2015 & Improved Vector Space Model & Information Retrieval & ZXing, SWT, and Guava

 \\ \hline
Almhana et al. \cite{almhana} & 2016 & NSGA-II & Search-based Software Engineering & Six open source datasets \cite{ye2014}
\\ \hline
Lam et al. \cite{lam2017} & 2017 & DNNLOC (rVSM + DNN) & Combining IR with DNN & Six Benchmark datasets \cite{ye2014}	
\\ \hline
Xiao et al. \cite{xiao2017} & 2017 & DeepLocator: Enchaned CNN using rTF-IDuF(Revised term frequency-user focused inverse document frequency) & Machine Learning & AspectJ, Eclipse, JDT, SWT and Tomcat projects
\\ \hline
Xiao et al. \cite{xiao2018} & 2018 &BugTranslator: Attention-based RNN Encoder-Decoder with LSTM cells & Deep Learning & Three open-source Java projects (Eclipse UI, JDT, and SWT)
\\ \hline

Huo et al. \cite{Huo2019} & 2019 & deep transfer learning approach named TRANP-CNN  & Deep Learning (CNN) & \\ \hline
Li et al. \cite{li2021} & 2021 & LaProb (Label Propagation-based software) & Semi-Supervised Machine Learning & Four Eclipse open source datasets \cite{zhou2012} and five open source datasets \cite{lee2018}
 \\ \hline

Fang et al. \cite{fang2021} & 2021 & LSTM, CNN, Multilayer perceptron & Deep learning & Four datasets \cite{ye2014}
\\ \hline

Zhu et al. \cite{Zhu2021} & 2021 & deep multimodal model, DEMOB and MDCcL encoder & LSTM and Deep CNN & Four Eclipse open source datasets \cite{lam2017}\\ \hline
Murali et al. \cite{Murali2021} & 2021 &Bug2Commit  & Word Embedding, BM25 & Real world dataset of Facebook application\\ \hline
Ciborowska et al. \cite{cibr2022} & 2022 &  Fast Bug Localization (FBL-BERT)& BERT &Six open source datasets \cite{ye2014} \\ \hline
Mahajan et al. \cite{mahajan2022} & 2022 &CNN optimized using hybridized cuckoo search-based sea lion optimization (CS-SLnO) & Improved Deep Learning & Aspect J and SWT
  \\ \hline
Hossain et al. \cite{hossain2024} & 2024 & Empirical Study of LLM-based Bug Localization and Repair & Large Language Models (Transformer-based NLP) & Multiple open-source software projects used for empirical evaluation
\\ \hline

Hu et al. \cite{hu2024multiobj} & 2024 & Multi-objective feature fusion-based fault localization framework & Multi-objective Optimization with Deep Learning & Public fault localization benchmark datasets
\\ \hline
Samir et al. \cite{samir2025brain} & 2025 & BRaIn (IR-based Bug Localization with Intelligent Relevance Feedback) & Information Retrieval with LLM-based Feedback & Multiple open-source Java projects evaluated at ICPC 2025
\\ \hline

\end{tabular}
\label{table2.1}
\end{table*}

\end{center}
\section{Literature Review} \label{literature}

Recent research on bug localization can be broadly classified into two main categories: bug localization using Information Retrieval (IR),deep learning and large language models, and bug localization using computational search-based approaches. Table \ref{table2.1} summarizes the most relevant studies in this area.

\subsection{Bug Localization using Information Retrieval, Deep Learning and Large Language Models}
Bug localization based on IR techniques aims to identify semantic and textual similarities between bug reports and source code artifacts. Classical approaches employ models such as Latent Dirichlet Allocation (LDA) \cite{blei2003}, Latent Semantic Indexing (LSI) \cite{dumais2004}, and the Vector Space Model (VSM) \cite{salton1975}.

BugLocator \cite{zhou2012} utilizes VSM to identify suspicious buggy classes corresponding to a given bug report. BugScout \cite{nguyen2011} applies a topic-based model using LDA to trace bug origins. BLUiR \cite{saha2013} decomposes bug reports into structured components and applies structural retrieval to rank candidate files. Rahman et al. \cite{rahman2015} propose an enhanced vector space model that integrates structural IR and version history, achieving improved performance on datasets such as SWT, ZXing, and Guava. Lam et al. \cite{lam2017} address lexical mismatches by combining IR techniques with Deep Neural Networks (DNN), introducing the DNNLOC model that employs autoencoders and deep learning to bridge semantic gaps between bug reports and source code.

Recent advancements increasingly leverage deep learning architectures for bug localization. Huo et al. proposed TRANP-CNN, a transfer learning-based convolutional neural network that improves cross-project bug localization using labeled target-project data. Zhu et al. introduced DEMOB, a deep multimodal framework that integrates deep convolutional neural networks and bidirectional long short-term memory (BiLSTM) models to capture multi-grained structural features. Ciborowska et al. proposed FBL-BERT, a BERT-based approach optimized for fine-grained bug localization through changeset-based matching. Murali et al. presented Bug2Commit, which applies vector space models, word embeddings, and BM25 to an industrial environment, although its performance depends on lexical overlap. Akbar et al. conducted a large-scale comparative study, demonstrating the superiority of third-generation tools, particularly those leveraging word embeddings for cross-language retrieval.

In addition, Hossain et al. presented a comprehensive empirical study on the application of large language models for automated bug localization and repair, highlighting their strengths, limitations, and practical challenges when compared to conventional approaches \cite{hossain2024}. Samir and Rahman introduced BRaIn, an improved IR-based bug localization technique that incorporates intelligent relevance feedback generated by large language models to reformulate queries and re-rank results, leading to consistent improvements in localization accuracy \cite{samir2025brain}.

\subsection{Bug Localization using Computational Search Approach}

Although IR-based techniques are widely adopted, they often suffer from false positives. Fang et al. \cite{fang2021} introduced a classification-based approach to distinguish informative from uninformative bug reports, demonstrating that LSTM models outperform SVM, CNN, and multilayer perceptron classifiers when precision is prioritized.

Pandiyan et al. \cite{pandiyan2015} proposed Tabu Search Fault Localization with Path Branch and Bound (TSFL-PBB), employing a two-phase approach to efficiently identify buggy code segments. Almhana et al. \cite{almhana} formulated bug localization as a multi-objective optimization problem using NSGA-II and evaluated the approach on six open-source datasets \cite{ye2014}. Xiao et al. \cite{xiao2017} introduced DeepLocator, combining CNNs with rTF-IDuF and word2vec embeddings to improve localization accuracy. BugTranslator \cite{xiao2018} further addressed lexical gaps through deep semantic translation between bug reports and source files. Li et al. \cite{li2021} proposed LaProb, a label propagation-based approach that models bug localization as a multi-label distribution learning problem using a Biparty Hybrid Graph. Mahajan et al. \cite{mahajan2022} introduced a hybrid Cuckoo Search–Sea Lion Optimization framework for feature extraction and classification, reporting improved results over traditional deep learning and IR-based models. Recent work has also explored optimization-based and hybrid approaches to bug localization. Hu et al. proposed a multi-objective deep learning framework that combines fault-proneness features with semantic information to improve fault localization accuracy.\cite{hu2024multiobj}.

In summary, the literature highlights bug localization as a critical challenge in software development and maintenance. While IR-based approaches provide a solid foundation, they remain susceptible to lexical mismatches and false positives. Recent research increasingly incorporates deep learning, evolutionary optimization, and hybrid techniques to enhance localization accuracy. Despite significant progress, there remains considerable scope for improvement, particularly in balancing relevance and result size. Motivated by these findings, this study advances bug localization by enhancing the use of multi-objective evolutionary algorithms within a unified optimization framework.

\section{Multi-Objective Problem Formulation for Bug Localization}
\label{methodology}
In this section, we provide an overview of our multi-objective approach for identifying and prioritizing significant code segments in bug reports. We then delve into the components of our proposed multi-objective solution formulation.

\subsection{Model Overview}

Our approach for identifying relevant classes involves exploring a wide search space of source project classes based on the given bug report description. The search space is determined by the number of possible combinations of ranked classes, and the ranking of these classes is crucial. Since there can be multiple classes responsible for an error, it's imperative to inspect all the implicated buggy files to mitigate the bugs.

Given the vast search space, traditional trial-and-error and other random search-based techniques would be computationally expensive and time-consuming. Therefore, we employ metaheuristic techniques capable of efficiently exploring large search spaces and generating potential solutions within a reasonable time frame \cite{suddle2022}. Our optimization approach is heuristic-based and centers on two primary, contradictory objectives: a correction function and a minimization function. The correction function is further divided into two sub-operations: maximizing the lexical similarity between the bug report description and the recommended buggy files, and maximizing the history-based score \cite{sheikh2024}. This history-based score considers recent changes made by programmers in response to past bug reports, correspondences between the current bug report and recent bug reports, and the number of suggested classes that were previously corrected. The minimization function focuses on reducing the number of recommended files or classes. These two conflicting objectives are addressed using metaheuristic techniques such as NSGA-II with enhanced tokenization, SPEA-2, and MOEA/D.

\begin{figure*}[ht]
\centerline{\includegraphics[scale=0.6]{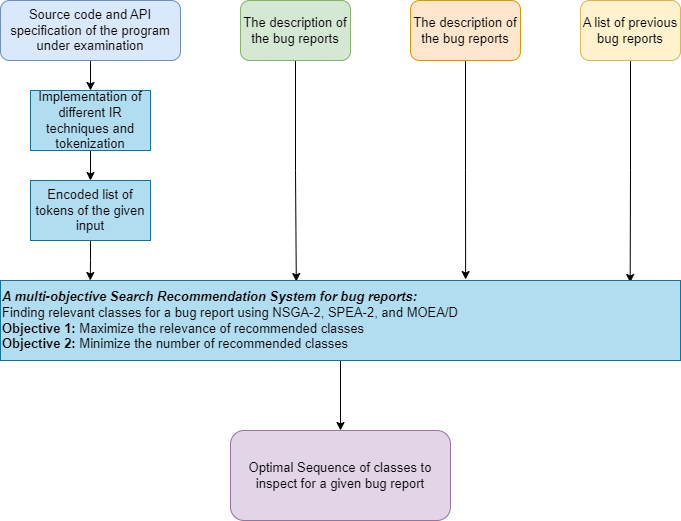}}
\caption{An overview of the proposed multi-objective Search Recommendation System for bug reports}
\label{fig:5}
\end{figure*}

Our model takes four parameters as input: the API specification of the class, a list of previous bug reports, the description of the current bug report, and the history of modifications in earlier software releases prompted by past bug reports. As output, our model generates an optimal list of recommended files or classes that closely match the bug report descriptions in terms of history-based and lexical similarity scores. Figure \ref{fig:5} provides an overview of our proposed model. In the subsequent sections, we begin by presenting a detailed overview of the three evolutionary algorithms (MOEA/D and SPEA-2, NSGA-II) utilized in this research. Following that, we outline the objectives or fitness function employed in our study. Subsequently, we elucidate the opposition-based learning strategy integrated into the evolutionary algorithms, specifically MOEA/D and SPEA-2.

\subsection{Strength Pareto Evolutionary Algorithm (SPEA-2) using opposition-based learning}
SPEA-2, an extension of the Strength Pareto Evolutionary Algorithm \cite{zitzler2001}, is designed to determine the dominance rank of each solution and evaluate their strength. The core objective of SPEA-2 is to identify and preserve a front that encompasses all non-dominated solutions, referred to as Pareto Optimal solutions, through evolutionary methods. These methods involve exploring the search space and generating child solutions using crossover and mutation operations. To pinpoint the non-dominated front, strength is determined based on the dominance rank and the density estimation of the Pareto optimal front \cite{rivas2005}.

\begin{figure}[ht]
\centerline{\includegraphics[width=80mm,height=70mm,scale=1]{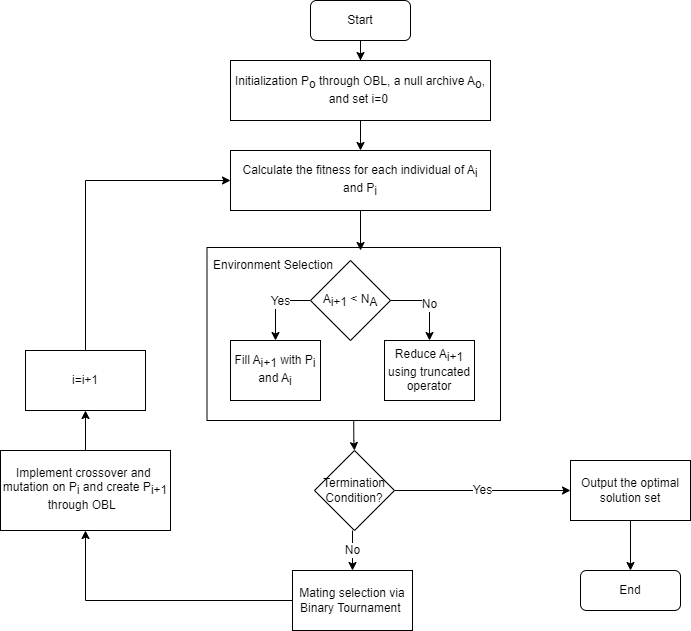}}
\caption{Flowchart of proposed SPEA-2 through opposition based learning}
\label{fig:4}
\end{figure}

SPEA-2 employs an external archive that contains the previous non-dominated solutions and updates it with each generation. Each solution is assigned a strength value, which is used to calculate its fitness \cite{gadhvi2016}. The flowchart depicting SPEA-2 with opposition-based learning is illustrated in Figure \ref{fig:4}.

The basic steps for SPEA-2 are as follows \cite{rivas2005}:
\begin{itemize}
    \item \textbf{Initialization:} Create an initial population $P_o$ and generate a null archive $A_o$ while setting i=0
\item \textbf{Fitness Calculation} Measure the fitness value of all individuals in $P_i$ and $A_i$.
\item \textbf{Environment Selection} Duplicate all the non-dominated individual from $P_i$ and $A_i$ in $A_{i+1}$. If the size($A_{i+1}$) < $N_A$ (archive size) then copy all individuals from $P_i$ and $A_i$ in $A_{i+1}$ otherwise reduce $A_{i+1}$ using a truncation operator.
 \item \textbf{Termination Condition} If the stopping criteria are met then set A to the set of non-dominated individuals in $A_{i+1}$. Halt
\item \textbf{Mating} To fill the mating pool, implement binary tournament selection with a replacement on $A_{i+1}$.
\item \textbf{Variation} Implement recombination and mutation to the mating pool and create an opposite population from the resulting population. Set $P_{i+1}$ from the population generated through opposition-based learning. Increment the counter $i=i+1$ and jump to the fitness calculation step.
\end{itemize}

\subsection{Multi Objective Evolutionary Algorithms using Decomposition Approach (MOEA/D)}
Multi-objective optimization problems can be effectively addressed through the decomposition approach introduced by Zhang et al. \cite{zhang2007}. This method decomposes a multi-objective optimization problem into several sub-problems using aggregation functions and solves them simultaneously with different evolutionary algorithms. Several decomposition techniques exist, including penalty-based boundary intersection, the Tchebycheff approach, and the weighted sum approach \cite{zheng2018}. Zhang et al. \cite{zhang2007} demonstrated that this strategy outperforms other state-of-the-art Multi-Objective Evolutionary Algorithms (MOEAs).

If the optimal solutions of each sub-problem are verified as Pareto optimal for the overall multi-objective problem, then the combined set of these solutions provides a strong approximation of the Pareto front. In our model, MOEA/D is implemented using opposition-based learning. Key features of MOEA/D include the following \cite{zheng2018}:
\begin{itemize}
\item MOEA/D does not directly solve a multi-objective problem; it optimizes N scalar problems simultaneously, making it compatible with other Evolutionary Algorithms (EAs).
\item It offers a lower computational complexity compared to NSGA-II and performs admirably in terms of complexity and solution quality.
\item The performance of MOEA/D can be enhanced with the utilization of advanced decomposition techniques.
\end{itemize}

\subsection{Non-dominated Sorting Genetic Algorithm (NSGA-II) }\label{token}
NSGA-II, introduced by Deb et al. \cite{deb2000}, is designed to identify a set of optimal solutions known as the Pareto set. This set represents a balanced compromise among conflicting objectives without sacrificing any of them. Each candidate solution is represented as a vector, where each dimension corresponds to a potential buggy class recommended for a given bug report.

\subsection{New Tokenization Approach}
In a previous study by Almhana et al. \cite{almhana}, NSGA-II was combined with a Camel-case Splitter-based tokenization technique to compare class names with a specific bug report. In our research, we adopt a different tokenization strategy that focuses only on essential elements of a class; such as comments, class names, function names, and variable names , rather than applying camel case splitting to the entire codebase. This approach targets classes that are more likely to be responsible for the bug while excluding irrelevant files, such as database connection or modeling code. Tokenizing non-relevant code increases storage requirements and negatively affects bug localization computation time. The results of this improved preprocessing technique using NSGA-II are presented in Tables \ref{table5.1}, \ref{table5.2}, and \ref{table5.3}, as well as in Figure \ref{fig:6}.

\subsection{Fitness Function (Objectives)}
The fitness function consists of two sub-functions (objectives): the correction function and the minimization function.
\subsubsection{First Objective: Increase Similarity}
The correction function (first objective) is a combination of lexical-based similarity (LS) and history-based similarity (HS). Formally, the correction function can be defined as follows:
\begin{equation}
f_1 = \frac{LS + HS}{2}
\end{equation}

\begin{figure}[ht]
\includegraphics[height = 0.35\textwidth]{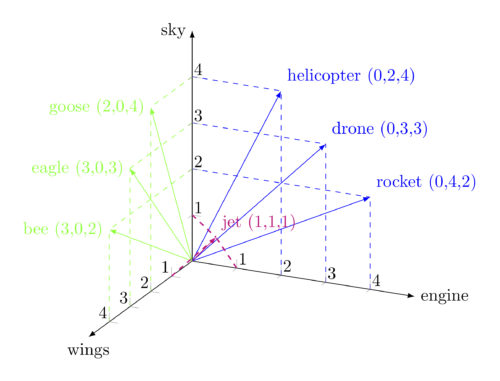}
\caption{An example of vectors space model for text, each vocabulary word represents a dimension}
\label{fig:cosine}
\end{figure}

Lexical similarity is determined by computing two functions. The first function calculates the cosine similarity \cite{wiki} between the vector representation of source code and the bug report description. The initial step in transforming text data into vectors involves preprocessing. This process includes tokenization, which entails splitting words, as well as the removal of stop words and stemming. For handling identifiers within the source code, a Camel case splitter is employed for effective tokenization. A standardized set of information retrieval stop words is utilized to eliminate irrelevant terms. Additionally, a Porter stemmer is applied to ensure consistency between related words, such as "exercising" and "exercised." Following this preprocessing, the text is converted into vectors using a Bag of Words (BOW) model. In this model, each dimension of the vector corresponds to a vocabulary term, and cosine similarity is then computed between two vectors. The orientation of a vector is determined by the word distribution in the text from which the vector is derived. If a word is more frequently present in a text, the corresponding vector will exhibit greater magnitude in the direction of that word. The cosine of the angle between two vectors is 0 when they are perpendicular (90 degrees angle). Two pieces of text will have a 90-degree angle between them if they possess significantly different word distributions. Conversely, if two vectors are very similar, the angle between them approaches zero, and the cosine of 0 is 1. Therefore, cosine similarity ranges from 0 (indicating dissimilarity) to 1 (indicating similarity).

Figure \ref{fig:cosine} shows an example of vector space model. The conceptual similarity between two vectors, denoted as $c_1$ and $c_2$, is assessed as follows:
\begin{equation}
\begin{split}
Sim(c_1,c_2) & = Cos(\Vec{c_1},\Vec{c_2}) = \frac{\Vec{c_1}.\Vec{c_2}}{\vert \vert \Vec{c_1}\vert \vert \times \vert \vert\Vec{c_2} \vert \vert } \\
& = \frac{\sum_{i=1}^{n}(w_{i,1}\times w_{i,2})}{\sqrt{\sum_{i=1}^{n}(w_{i,1})^{2} \times \sum_{i=1}^{n}(w_{i,2})^2}} \in [0,1]
\end{split}    
\end{equation}
Where $\Vec{c_1} = (w_{1,1},w_{2,1},\ldots w_{n,1})$  is the term vectors that corresponds to $c_1$. The weights $w_{i,j}$ are calculated using IR strategies such as term frequency–inverse term frequency \cite{almhana}.    

Considering that the bug report is composed in natural language, while the code is written in a programming language, it's evident that the cosine similarity would be higher if the code elements have detailed comments or if the specific names of the problematic code elements are mentioned in the bug report. To account for this situation, another measure of lexical similarity is computed between the bug report description and the API documentation. Another element of the correction function involves the history-based similarity, which encompasses three sub-functions. The first sub-function addresses the observation that recently fixed code snippets have a higher likelihood of causing a bug again. To mitigate this issue, this sub-function leverages the history of bug reports to track how many times a specific code snippet has been fixed.

\begin{equation}
    H_1 = \frac{\sum_{i=1}^{Size(S)} NbFixedBugs(report, C_i)}{Size(S) \times Max(NbFixedBugs(report,C_i))} \in [0,1]
\end{equation}

Here S is the solution consisting of a number of recommended classes $S={c_1,c_2,...,c_{size(S)}}$ \cite{almhana}. Another observation is that a code that is rectified earlier has a higher chance of causing an error. Therefore, the second function compares the last date on which the source code was modified and the date of the bug report.   
\begin{equation}
    H_2 = \frac{\sum_{i=1}^{Size(S)} \frac{1}{reportdata - last(report, c_i)+1}}{Size(S)} \in [0,1]
\end{equation}

The last function measures the consistency among the recently occurred bug reports and the recommended source codes. Equation \ref{fun3} figures the cardinality, Crd, of the sets of classes recommended together for earlier bug reports and the largest conjunction set of files/classes between the solution S. 
\begin{equation} \label{fun3}
    H_3= \frac{Crd}{Size(S)} \in [0,1]
\end{equation}
\subsubsection{Second Objective: Minimize number of classes}
The minimization function (second objective) aims to reduce the number of suggested buggy classes. This objective evaluates the solution size for a given bug report and ensures that only a minimal set of classes is recommended for each case.

These two objectives are inherently contradictory. Increasing the similarity threshold to improve class ranking raises the likelihood of retrieving more similar classes (supporting the first objective). However, doing so negatively affects the second objective by increasing the number of retrieved classes.

\subsection{Opposition-based Learning (OBL)}
In multi-objective algorithms, the optimization process begins with an initial population and evolves it across generations to approach the Pareto-optimal solution. This iterative procedure continues until the predefined stopping criteria are satisfied.

Opposition-based learning \cite{tizhoosh2005} can be effectively applied during both the initialization and evolution phases. Inspired by the ancient yin–yang philosophy, it emphasizes the complementary nature of opposites \cite{iacca2011}. In practice, this technique generates opposite solutions during initialization and evolution rather than relying solely on random numbers, thereby accelerating convergence.

Rahnamayan et al. demonstrated that independent random numbers can be up to 50\% farther from the best solution compared to their opposite counterparts \cite{rahnamayan2008}. This finding highlights the effectiveness of opposition-based learning in enhancing the optimization process.

Let $ x \in \Re$ a real number lies in the interval $x \in [a,b]$ \cite{tizhoosh2005}. The opposite number $ \tilde{x} $ is defined as:

\begin{equation}
    \tilde{x}  = a+b-x 
\end{equation} 

Consider the function $f(x)$ and the evaluation function $g(\cdot)$. Let $x$ be an independent random guess, sampled from the interval $[a,b]$, and let $\Tilde{x}$ represent its opposite value. After each generation, we compute the values of $f(x)$ and $f(\Tilde{x})$. The learning process continues with $\Tilde{x}$ if $g(f(x)) \leq g(f(\Tilde{x}))$; otherwise, it proceeds with $x$. It's important to note that the evaluation function $g(\cdot)$ serves as the yardstick for optimality, comparing results such as error function, fitness function, rewards, and penalties \cite{tizhoosh2005}. This approach helps in making decisions that lead to better solutions during the optimization process.

\begin{center}
\begin{table*}[t]
\footnotesize
\caption{Dataset Description}
\begin{tabular}{|p{2cm} |p{2cm}|p{2cm}|p{2cm}|p{1cm}|}
 \hline
Project Name & No. of Bug Reports & No. of files in project & No. of fixed classes per bug report & No. of API \\ [0.5ex] 
 \hline\hline
SWT & 4151 &2056 & 3 & 161\\ \hline
AspectJ & 593 &4439 & 2 & 54\\ \hline
Tomcat & 1056 &1552 & 1 & 389\\ \hline
Birt  & 4178 &6841 & 1 & 957\\ \hline
JDT & 6274 &8184 & 2 & 1329\\ \hline
Eclipse UI & 6495 &3464 & 2 & 1314  \\ 
[0.5ex] 
 \hline
\end{tabular}
\label{table3.1}
\end{table*}
\end{center}

\subsection{Dataset Description}
\begin{figure}[ht]
\centerline{\includegraphics[width=80mm,height=70mm,scale=1]{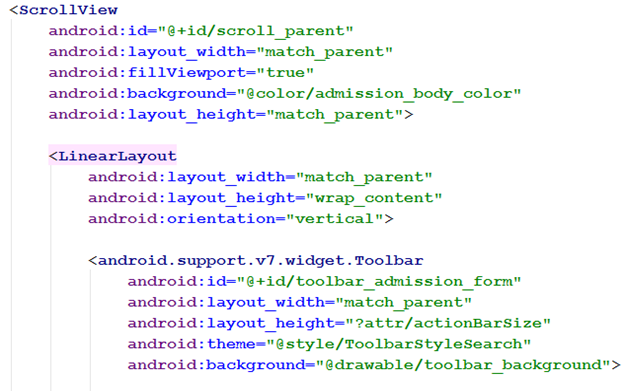}}
\caption{A code fragment written in XML programming language}
\label{fig:2}
\end{figure}

\begin{figure}[ht]
\centerline{\includegraphics[width=80mm,height=70mm,scale=1]{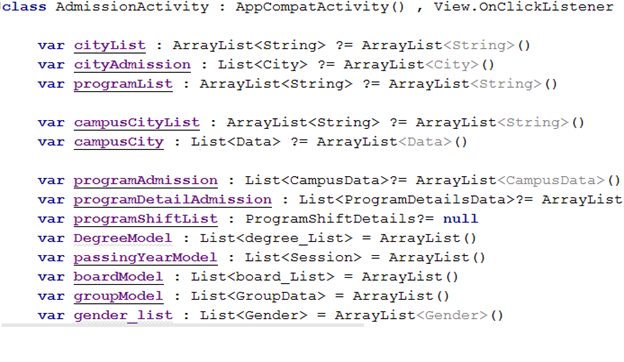}}
\caption{A code fragment from class AdmissionActivity written in Kotlin programming language}
\label{fig:3}
\end{figure}

To evaluate our proposed model, we conducted experiments on six open-source software projects comprising more than 22,000 bug reports. The dataset attributes are summarized in Table \ref{table3.1}. For additional validation, we also applied our model to a non-open-source Kotlin-based industrial Android dataset containing 100 bug reports. During experimentation with the Android project, we encountered two key challenges: constructing a bug report database and performing data parsing. To validate the model’s output, we created a database linking buggy classes to their corresponding bug reports. While the previous projects were written entirely in Java—making lexical token parsing, the Android project required additional effort due to its heterogeneous structure. The project includes three types of files: Kotlin, Java, and XML. Kotlin is a statically typed programming language that is relatively new to Android development, while XML files contain the front-end components of the application and are linked to backend files written in Kotlin or Java. Examples of XML and Kotlin code are shown in Figure \ref{fig:2} and Figure \ref{fig:3}, respectively.

\section{Evaluation Techniques}
\label{eval}
We developed a set of evaluation techniques to rigorously assess the validity of our proposed framework. Our experiments were conducted on a large collection of Java-based open-source projects as well as an Android project. To ensure robustness, each experiment was replicated 30 times. We also performed a comprehensive comparison between our model and previously published heuristic and non-heuristic approaches implemented on the same datasets. The credibility of our model is examined through four research questions, with results discussed accordingly.

\subsection{Research Questions (RQ)}
We evaluate the performance of our proposed system by addressing the following research questions (RQ) and analyzing the results accordingly.

\begin{enumerate}
\item RQ1: Validation of Efficacy — To what extent does the proposed framework effectively identify relevant classes based on bug reports?

\item RQ2: Validation of Improved Tokenization Method — How does our model's performance using the improved tokenization technique for NSGA-II compare to standard NSGA-II?

\item RQ3: Comparison with Mono-Objective Heuristics Approaches — How does our model perform relative to single-objective heuristic approaches?

\item RQ4: Cross-Domain Validation — Can our proposed model demonstrate strong performance when evaluated on bug reports from software developed in different programming languages?
\end{enumerate}

These research questions provide a structured basis for evaluation and help us draw meaningful conclusions regarding the performance and effectiveness of the proposed system.

To address RQ1, we evaluate the multi-objective models on multiple open-source datasets to assess the list of recommended classes responsible for each bug. We consider three evaluation metrics: 
\begin{itemize}
\item \textbf{Precision@k}   For a given integer `k', the precision is the number of accurately suggested classes in the top `k' of suggested classes divided by the lowest number of classes to examine in the ranked suggestions list. It is calculated as:
\begin{equation}
    p=\frac{t_p}{(t_p + f_p)}
\end{equation}
Here $f_p$ is the number of false positive results, and $t_p$ is the number of true positive results.
\item \textbf{Recall@k}   For a given integer 'k', the recall is the number of correct suggested classes in the top k of suggested classes by the solution divided by the total number of expected classes to be suggested that possess an error. It is calculated as:
\begin{equation}
    r=\frac{t_p}{(t_p + f_n)}
\end{equation}
Here $t_p$ is the number of true positive results, and $f_n$ is the number of false negative results.

\item \textbf{Accuracy@k}  For a given integer 'k', the accuracy is the ratio of those bug reports against which the recommended classes had at least one valid reading within the top 'k' recommended classes.
\end{itemize}

To address RQ2, we introduced an improved tokenization technique for NSGA-II. Instead of tokenizing the entire source code, we tokenize only the relevant components. We executed the model using both standard NSGA-II (with full-code tokenization) and NSGA-II with improved tokenization. A detailed comparison of the results is presented in the following section.

For RQ3, we compared our results with those of a mono-objective optimization model. Comparing a mono-objective model with a multi-objective model is challenging because the former yields a single optimal solution, whereas the latter produces a set of non-dominated solutions. To enable comparison, we selected the solution closest to the Knee point as the representative candidate for the multi-objective model.

To address RQ4, we constructed a bug report database for an Android project containing XML, Kotlin, and Java files. The repository includes one column for bug reports and another for the corresponding buggy classes. To validate our model’s accuracy, we executed it on this industrial Android project.

\section{Results and Discussion} \label{result}
\subsection{Results for RQ1}
\begin{figure}[ht]
\centerline{\includegraphics[width=80mm,height=50mm,scale=1]{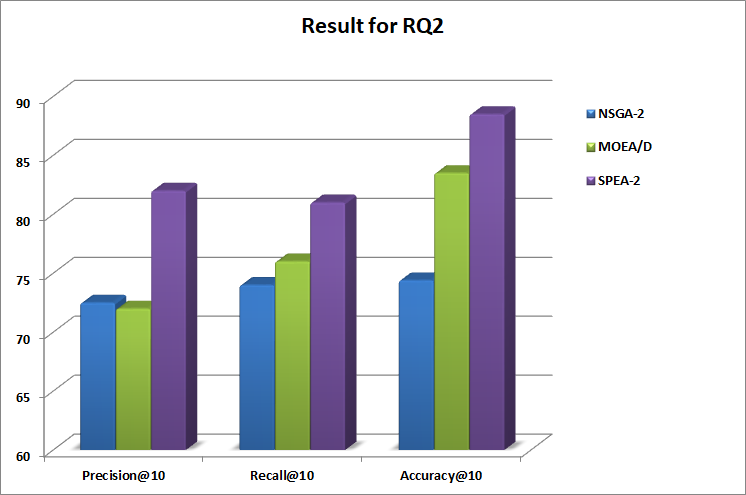}}
\caption{Average Precision@10, Recall@10, and Accuracy@10 for NSGA-II, MOEA/D, and SPEA-2 on six open source java datasets listed in Table \ref{table3.1}}
\label{fig:7}
\end{figure}

To demonstrate the effectiveness of our proposed model, we recorded precision, recall, and accuracy at various values of k ranging from 5 to 20. The last two columns of Tables \ref{table5.1}, \ref{table5.2}, and \ref{table5.3} present the evaluation metric values for six open-source Java projects. Figure \ref{fig:7} shows that SPEA-2 outperforms MOEA/D and NSGA-II, particularly in terms of accuracy at k=10. SPEA-2 achieved average precision@k values of 86\%, 82\%, 76\%, and 71\%, while MOEA/D produced 75\%, 72\%, 69\%, and 64\%, and NSGA-II yielded 83\%, 80\%, 71\%, and 68\%. SPEA-2 consistently records the highest precision, with similar trends observed for recall and accuracy.

At k=20, recall is highest for all models and decreases as k decreases. For SPEA-2, accuracy reaches 94\% at k=20, indicating that 94\% of the correct buggy classes were identified across the numerous bug reports in each software project.

Regarding RQ2, the results obtained using various evaluation metrics clearly support our hypothesis that the proposed models can effectively identify buggy classes for each bug report.

\subsection{Results for RQ2}
\begin{figure}[ht]
\centerline{\includegraphics[width=80mm,height=50mm,scale=1]{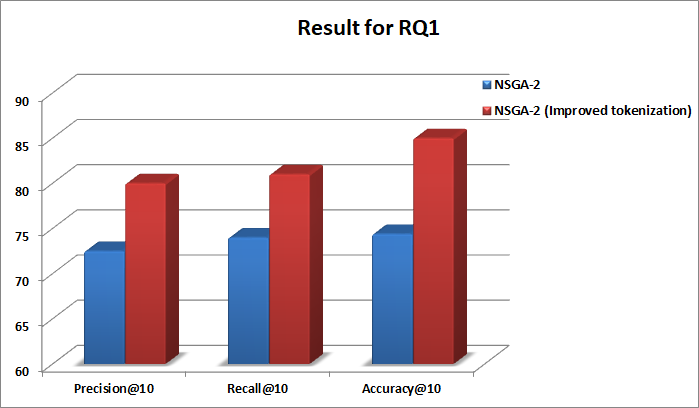}}
\caption{Average Precision@10, Recall@10, and Accuracy@10 for NSGA-II and NSGA-II(Improved tokenization) on six open source java datasets listed in Table \ref{table3.1}}
\label{fig:6}
\end{figure}

The results presented in Figure \ref{fig:6} and Tables \ref{table5.1}, \ref{table5.2}, and \ref{table5.3} highlight the effectiveness of the improved preprocessing technique applied before using NSGA-II. The second and third columns of these tables show the average precision, recall, and accuracy at k=5, k=10, k=15, and k=20. The most notable difference in precision appears at k=10, where the lowest precision for NSGA-II (improved) is 68\%. This reduction is acceptable given the lower generalization level at k=20. In terms of recall, an average of 71\% of the predicted buggy classes were identified within the top 5 rankings. Table \ref{table5.2} further confirms that most actual buggy classes appear within the top 20, with an average recall of 90\%.

We implemented NSGA-II using improved tokenization (Column 3), which considers only the critical sections of the source code, whereas standard NSGA-II (Column 2) tokenizes the entire codebase. Our proposed method also outperforms standard NSGA-II in terms of accuracy, achieving 95\% accuracy at k=20. These results clearly demonstrate that focusing on the critical components of the code not only reduces computational time and storage requirements but also enhances overall model performance.

\begin{center}
\begin{table*}[t]
\footnotesize
\caption{Precision@k}
\begin{tabular}{|p{0.5cm}|p{1.1cm}|p{2.5cm}|p{1.2cm}|p{1.3cm}|p{1.2cm}|p{1cm}|}
 \hline
k & NSGA-II & NSGA-II using proposed tokenization & Lexical similarity & History-based Similarity & MOEA/D & SPEA-2 \\ [0.5ex] 
 \hline\hline
5 & 76 &83 & 67 & 68 & 75& \textbf{86} \\ \hline
10 & 72.5 &80 & 60 & 60 & 72& \textbf{82}\\ \hline
15 & 65 &71 & 57 & 51 & 69& \textbf{76}\\ \hline
20  & 61.4 &68 & 48 & 47 &64 & \textbf{71}\\ 
[0.5ex] 
 \hline
\end{tabular}
\label{table5.1}
\end{table*}
\end{center}

\begin{center}
\begin{table*}[t]
\footnotesize
\caption{Recall@k}
\begin{tabular}{|p{0.5cm}|p{1.1cm}|p{2.5cm}|p{1.2cm}|p{1.3cm}|p{1.2cm}|p{1cm}|}
 \hline
k & NSGA-II & NSGA-II using proposed tokenization  & Lexical similarity & History-based Similarity & MOEA/D & SPEA-2 \\ [0.5ex] 
 \hline\hline
5 & 67 &71 & 53 & 61 & 69.5&\textbf{73} \\ \hline
10 & 74 &\textbf{81} & 43 & 57 & 76& \textbf{81}\\ \hline
15 & 79.5 &86 & 57 & 63 & 81.4& \textbf{86.5}\\ \hline
20  & 83 &90 & 71 & 68 &87 & \textbf{91.3}\\ 
[0.5ex] 
 \hline
\end{tabular}
\label{table5.2}
\end{table*}
\end{center}

\begin{center}
\begin{table*}[t]
\footnotesize
\caption{Accuracy@k}
\begin{tabular}{|p{0.5cm}|p{1.1cm}|p{2.5cm}|p{1.2cm}|p{1.3cm}|p{1.2cm}|p{1cm}|}
 \hline
k & NSGA-II & NSGA-II using proposed tokenization  & Lexical similarity & History-based Similarity & MOEA/D & SPEA-2 \\ [0.5ex] 
 \hline\hline
5 & 61 &65 & 49 & 32 & 62&\textbf{68} \\ \hline
10 & 74.4 &85 & 72 & 63 & 83.5& \textbf{88.5}\\ \hline
15 & 82 &\textbf{92} & 81 & 70 & 90.8& 91\\ \hline
20  & 87.3 &\textbf{95} & 86 & 69 &92 & 94\\ 
[0.5ex] 
 \hline
\end{tabular}
\label{table5.3}
\end{table*}
\end{center}

\begin{center}
\begin{table*}[t]
\footnotesize
\centering
\caption{Results on Industrial Android project}
\begin{tabular}{|p{0.4cm}|p{0.8cm}|p{0.8cm}|p{0.8cm}|p{0.8cm}|p{0.8cm}|p{0.8cm}|p{0.8cm}|p{0.8cm}|p{0.8cm}|}
 \hline
 k & \multicolumn{3}{|p{0.9cm}|}{\textbf{Precision }}& \multicolumn{3}{|p{0.9cm}|}{\textbf{Recall}} & \multicolumn{3}{|p{0.9cm}|}{\textbf{Accuracy}}\\ \hline
& NSGA-II & MOEA /D& SPEA-2&NSGA-II & MOEA /D& SPEA-2&NSGA-II & MOEA /D& SPEA-2 \\ [0.5ex]
 \hline\hline
5 & 42.9 &39&\textbf{44}&18.8& 17&\textbf{20.4}&42& 39&\textbf{45}\\ \hline 
10 &50 & 47  &\textbf{53.5} & 43.8 &41.3  &\textbf{46.2} & 64  &62.5  &\textbf{67}  \\ \hline
15  & 52.7 &49.5  &\textbf{56}  &62.5  &59 &\textbf{65} &72  &67 &\textbf{74}\\ \hline
20  &44.5 &41  &\textbf{48} & 75 &71.8 &\textbf{78.2} &84 &75.3 &\textbf{88.5}\\
[0.5ex] 
 \hline
\end{tabular}
\label{table5.4}
\end{table*}
\end{center}
\subsection{Results for RQ3}
\begin{figure}[ht]
\centerline{\includegraphics[width=80mm,height=50mm,scale=1]{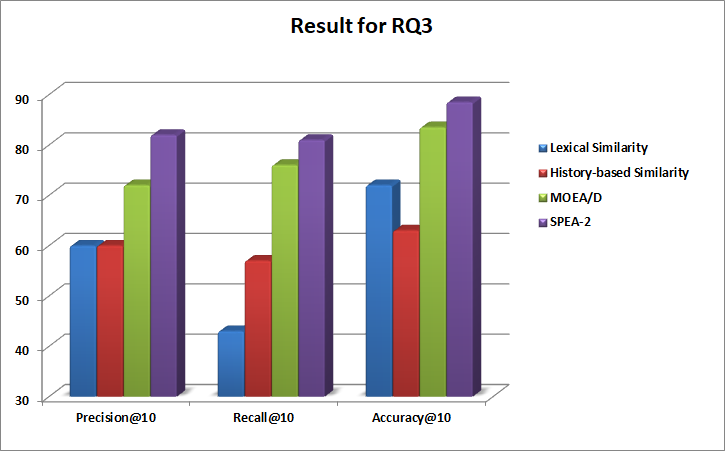}}
\caption{Precision@10, Recall@10, and Accuracy@10 for Lexical similarity, history-based similarity, MOEA/D, and SPEA-2 on six open source java datasets listed in Table \ref{table3.1}}
\label{fig:8}
\end{figure}

Regarding RQ3, we implemented two mono-objective formulations across all six projects and evaluated them using precision, recall, and accuracy. The mono-objective formulations consisted of an algorithm using only lexical similarity (LS) and another using only history-based similarity (HS). Figure \ref{fig:8} and Tables \ref{table5.1}, \ref{table5.2}, and \ref{table5.3} present the results, showing that both mono-objective approaches were outperformed by our proposed multi-objective models.

The average precision for LS and HS is 58.5\% and 56.5\%, respectively—both lower than the precision achieved by all multi-objective models. The highest recall recorded for LS is 71\%, and for HS it is 68\%, whereas SPEA-2 achieves a recall of 91.3\% at k=20. In terms of accuracy, LS performs better than HS by a margin of 17\%, but both are surpassed by NSGA-II and SPEA-2 by a substantial margin.

These findings highlight that combining both objectives significantly improves performance across all six open-source datasets. We also observed that some buggy classes missed by one mono-objective algorithm were correctly identified by the other, further illustrating the benefit of aggregating both objectives within a multi-objective framework.

\subsection{Results for RQ4}
\begin{figure}[ht]
\centerline{\includegraphics[width=80mm,height=50mm,scale=1]{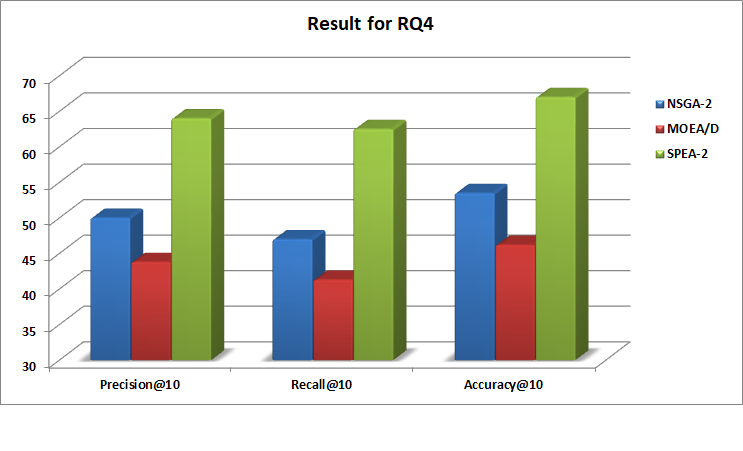}}
\caption{Precision@10, Recall@10, and Accuracy@10 for NSGA-II (Improved version), MOEA/D, and SPEA-2 on Kotlin-based Industrial Android project}
\label{fig:9}
\end{figure}

Thus far, the proposed multi-objective model has demonstrated strong performance across all six Java-based projects. To evaluate its cross-domain applicability, we tested our models on a Kotlin-based industrial Android project. Table \ref{table5.4} summarizes the results obtained using NSGA-II, MOEA/D, and SPEA-2, while Figure \ref{fig:9} presents a bar chart of the average precision, recall, and accuracy at k=10 for this dataset.

As noted earlier, the Android dataset is considerably smaller than the six Java datasets, which contributes to lower evaluation metric scores overall. Nevertheless, SPEA-2 outperforms NSGA-II and MOEA/D across all metrics. For precision, the highest score achieved by NSGA-II is 52.7\% at k=15, while MOEA/D records the lowest precision at every k value. In terms of recall, all models initially perform below average at k=5, but their performance improves as k increases. SPEA-2 achieves a recall of 78.2\% at k=20, indicating that 78.2\% of the correct buggy code elements were identified within the top 20. At k=20, SPEA-2 also records an accuracy of 88.5\%, compared to 84\% for NSGA-II and 75.3\% for MOEA/D.

These results confirm our hypothesis that the proposed model is independent of programming language and capable of delivering strong performance across different software development environments.

Table \ref{tableTime} presents an efficiency comparison, in milliseconds, between the proposed algorithms and existing similarity-based methods. The results clearly show that the proposed models achieve shorter running times compared to both lexical similarity and history-based similarity approaches.
\begin{center}
\begin{table*}[t]
\footnotesize
\caption{Efficiency of proposed algorithms (Running Time in milliseconds) }\label{tableTime}
\begin{tabular}{|p{2cm}|p{1.1cm}|p{1.1cm}|p{1.3cm}|p{2cm}|}
 \hline
NSGA-II (Improved version)  &  SPEA-2  & MOEA/D & Lexical similarity & History-based Similarity \\ 
 \hline
2874	&3410&	3123&	5982	&6863 \\ \hline
\end{tabular}

\end{table*}
\end{center}

\section{Conclusion and Future Work} \label{conclusion}
In this research paper, we introduced an automated multi-objective, class-level approach designed to localize and rank relevant buggy code elements in response to bug reports. Our proposed model balances maximizing the correctness function with minimizing the number of recommended buggy files. The correctness function is computed based on the similarity between bug reports, source code, and API documentation, as well as the history of bug-fixing activities and code changes prompted by previous bug reports. We evaluated our model on a small Kotlin-based Android dataset and six relatively large open-source Java datasets, collectively containing more than 22,000 bug reports.

Our search system demonstrated strong performance in identifying true buggy files, with nearly 94\% of error reports having the correct buggy files ranked within the top 20 suggested classes. Among the evaluated models, SPEA-2 achieved the highest precision and recall, outperforming NSGA-II, MOEA/D, and the mono-objective algorithms. These findings provide strong evidence of the efficiency and effectiveness of our multi-objective approach compared to both multi-objective and mono-objective baselines.

Looking ahead, we plan to further enhance our results by improving the preprocessing phase of the multi-objective model, with the goal of identifying an even larger proportion of buggy files within the top-ranked classes. Additionally, we aim to extend our research by applying the model to other programming platforms, such as Angular-JS and iOS, with the objective of identifying relevant code within the top five recommended classes. We also intend to explore the complexities associated with different types of errors when identifying relevant files, thereby further optimizing our search recommendation system.

\end{document}